\documentclass{article}

\usepackage[preprint]{neurips_2023}

\usepackage{natbib}
\setcitestyle{numbers,square}

\usepackage[utf8]{inputenc} 
\usepackage[T1]{fontenc}    
\usepackage{hyperref}       
\usepackage{url}            
\usepackage{booktabs}       
\usepackage{amsfonts}       
\usepackage{nicefrac}       
\usepackage{microtype}      
\usepackage{xcolor}         

\usepackage{graphicx}
\usepackage{subcaption}

\usepackage{algorithm}  
\usepackage{algpseudocode} 

\usepackage{comment}

\title{Integer or Floating Point? New Outlooks for\\ Low-Bit Quantization on Large Language Models}

\author{
  Yijia Zhang\thanks{These authors contributed equally to this work. Work done during internship at Microsoft Research Asia.} \\
  Shanghai Jiao Tong University\\
  \texttt{zhangyijia@sjtu.edu.cn} \\
  \And
  Lingran Zhao\footnotemark[1] \\
  Peking University \\
  \texttt{calvinzhao@pku.edu.cn} \\
  \AND
  Shijie Cao \\
  Microsoft Research Asia \\
  \texttt{shijiecao@microsoft.com} \\
  \And
  Wenqiang Wang \\
  Shanghai Jiao Tong University \\
  \texttt{wangwq20@sjtu.edu.cn} \\
  \And
  Ting Cao \\
  Microsoft Research Asia \\
  \texttt{ting.cao@microsoft.com} \\
  \And
  Fan Yang \\
  Microsoft Research Asia \\
  \texttt{fanyang@microsoft.com} \\
  \And
  Mao Yang \\
  Microsoft Research Asia \\
  \texttt{maoyang@microsoft.com} \\
  \And
  Shanghang Zhang \\
  Peking University \\
  \texttt{shanghang@pku.edu.cn} \\
  \And
  Ningyi Xu \\
  Shanghai Jiao Tong University \\
  \texttt{xuningyi@sjtu.edu.cn} \\
}

\begin{document}

\maketitle

\begin{abstract}

Efficient deployment of large language models (LLMs) necessitates low-bit quantization to minimize model size and inference cost.
While low-bit integer formats (e.g., INT8/INT4) have been the conventional choice, emerging low-bit floating-point formats (e.g., FP8/FP4) offer a compelling alternative and are gaining support from cutting-edge hardware, such as NVIDIA's H100 GPU. 
However, the superiority of low-bit INT versus FP formats for quantization on LLMs remains unclear.
In this study, we conduct a comparative analysis of INT and FP quantization with the same bit-width, revealing that the optimal quantization format varies across different layers due to the complexity and diversity of tensor distribution. 
Consequently, we advocate the \textbf{Mixture of Formats Quantization (MoFQ)}, which selects the optimal format on a layer-wise basis.
This simple yet effective approach achieves state-of-the-art results in both weight-only (W-only) and weight-activation (WA) post-training quantization scenarios when tested on LLaMA across various tasks.
In 4-bit W-only quantization, MoFQ surpasses GPTQ without complex hyperparameter tuning and with an order of magnitude faster quantization speed. 
While in 8-bit WA quantization, MoFQ significantly outperforms INT/FP-only methods, achieving performance close to the full precision model.
Notably, MoFQ incurs no hardware overhead compared to INT/FP-only quantization, as the bit-width remains unchanged.

\end{abstract}

\section{Introduction}

Low-bit quantization plays a crucial role in the deployment of deep learning models due to its capacity to minimize the model size and inference cost~\cite{polino2018model,han2015deep}. 
The recent advent of large language models (LLMs), which often contain tens or even hundreds of billions of parameters, has further intensified the need for effective quantization techniques~\cite{brown2020language,touvron2023llama}.
Conventional research in this domain has primarily focused on employing \textit{low-bit integer (INT)} formats for quantization~\cite{wu2020integer,lin2016fixed,jacob2018quantization}.
However, as the scale of LLMs expands, integer quantization faces challenges in maintaining the effectiveness observed in smaller models, thereby necessitating tailored optimizations or alternative approaches~\cite{dettmers2022llm,xiao2022smoothquant,frantar2022gptq}.

Recently, \textit{low-bit floating-point (FP)} formats have emerged as promising alternatives for DNN quantization~\cite{kuzmin2022fp8,micikevicius2022fp8,sun2019hybrid,agrawal20219}. 
FP8 has already garnered support from leading hardware vendors, such as NVIDIA, whose H100 GPU offers identical peak performance for FP8 and INT8 operations~\cite{h100}. Additionally, other CPU and GPU vendors, such as Intel, AMD, and Qualcomm, are actively incorporating FP8 capabilities into their hardware.
Compared to INT formats with the same bit width, FP formats offer a typically larger data range and higher precision for small values but have lower precision for large values and potentially higher hardware costs.

Given the notorious difficulty in quantizing LLMs, the relative effectiveness of INT and FP quantization remains ambiguous, motivating an interesting question:
\textit{Considering that INT and FP formats with the same bit width can represent the same number of discrete values (e.g., both INT8 and FP8 can represent $2^8=256$ values) but differ in value distribution, what are the distinct impacts on model quantization and efficient inference?}

To answer this question, we conduct a comparative analysis of INT and FP formats, focusing on hardware efficiency and quantization error.
First, we compare hardware efficiency by benchmarking the cost of FP and INT multiply-accumulate (MAC) units.
While FP MAC generally requires more hardware resources than INT MAC, the resource gap substantially narrows as the bit width decreases, with FP8 and INT8 costs being notably similar.
Next, we examine the quantization error of INT and FP formats through the statistical analysis of real tensors and layers sampled from the LLaMA model. Our findings reveal no consistent superior format for all tensors or layers due to the complexity and diversity of tensor distribution.
The optimal format is influenced by a combination of various factors, including the static/dynamic nature of tensors, outlier distribution, and quantization bit-width.
For weight tensors with static distribution, INT quantization outperforms FP quantization at 8-bit but this advantage diminishes at 4-bit.
For activation tensors with dynamic distribution and significant outliers, FP quantization surpasses INT quantization due to FP can represent large values with lower precision and small values with higher precision.

Inspired by the analysis finding of no consistent superior format, we propose the Mixture of Formats Quantization (MoFQ) approach, which selectively determines the optimal format from INT and FP with the same bit-width on a layer-wise basis. 
MoFQ effectively harnesses the complementary benefits of both formats, while ensuring that tensors within the same layer share the same data type and all quantized tensors have the same bit-width.
A straightforward format selection method proves effective, choosing the format with the minimum quantization error based on metrics such as tensor MSE, layer output MSE, or model output MSE.
MoFQ applies to W-only quantization for memory footprint compression and WA quantization for accelerated inference using lower-bit computation units.
For W-only quantization, MoFQ ensures compatibility with a broad range of existing hardware and imposes no additional hardware overhead compared to integer-only quantization, as the bit-width remains unaltered.
For WA quantization, MoFQ can be seamlessly integrated with hardware supporting both low-bit INT and FP operations, such as the off-the-shelf H100 GPU and upcoming Intel and AMD chips.
In summary, this mixed-format approach demonstrates simplicity, effectiveness, and efficiency in quantization format selection and model performance.

When evaluated on large language models across various tasks, our MoFQ achieves state-of-the-art (SOTA) results on both W-only quantization and WA quantization. 
For W-only quantization with 4-bit, MoFQ achieves comparable or better accuracy than GPTQ with an order of magnitude faster quantization speed, because GPTQ is based on second-order information and our MoFQ adopts the naive linear quantization with round-to-nearest (RTN).
For WA quantization with 8-bit, MoFQ significantly outperforms INT8-only and FP8-only quantization methods and achieves performance close to the full precision model.

Our contributions can be summarized as follows:
\begin{enumerate}
    \item We conduct an in-depth comparative analysis of INT and FP formats for quantizing LLMs, offering valuable insights to guide future quantization designs with low-bit FP formats.
    \item We propose the Mixture of Formats Quantization (MoFQ) approach, a simple yet effective layer-wise format selection scheme that seamlessly combines the benefits of both INT and FP formats. MoFQ is system and hardware-friendly.
    \item MoFQ achieves state-of-the-art (SOTA) results on both 4-bit W-only quantization and 8-bit WA quantization.
\end{enumerate}

\section{Preliminaries}

\paragraph{Integer vs. Floating Point Formats}

Integer and floating point formats are two primary ways to represent numbers in computing systems. The key distinction between them lies in the value distribution.
Integer format has a uniform distribution across the representable range with a difference of 1 between two consecutive numbers. While the floating point format exhibits a non-uniform distribution due to the incorporation of the exponent and mantissa design, thus providing higher precision for smaller numbers and lower precision for larger ones.
The emerging FP8 format features two widely-adopted options, E4M3 and E5M2, with fewer bits for both exponent and mantissa than FP32/16, as depicted in Figure~\ref{fig:fp_format}.
To highlight FP and INT distribution differences, Figure~\ref{fig:distribution} visualizes a small portion of values around the zero point represented in INT8 and FP8-E5M2.

two figures: 

\begin{figure}[!htb]  
  \centering  
  \begin{minipage}{0.4\textwidth}  
    \centering  
    \includegraphics[width=\linewidth]{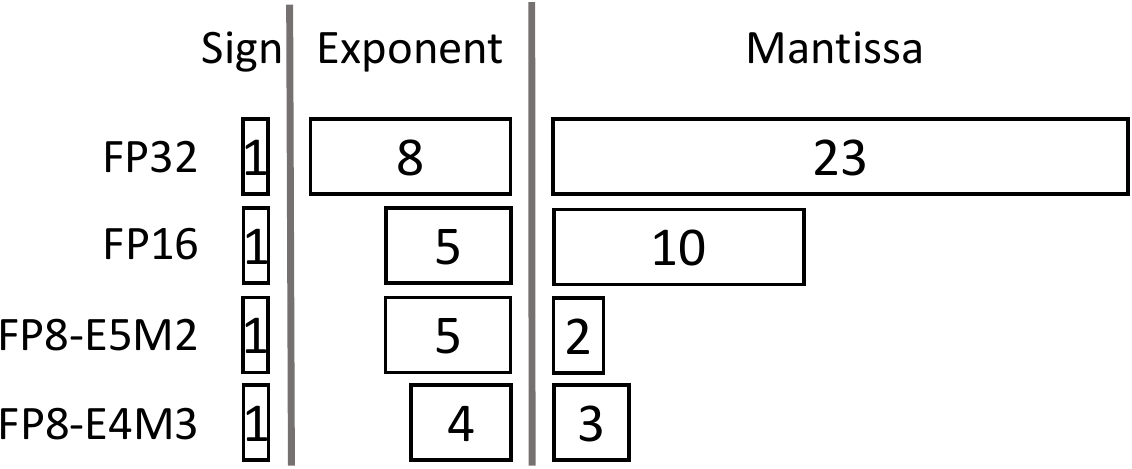} 
    \vspace{0.05in} 
    \caption{Structure of FP formats.}  
    \label{fig:fp_format}  
  \end{minipage}  
  \hfill  
  \begin{minipage}{0.58\textwidth}  
    \centering  
    \includegraphics[width=\linewidth]{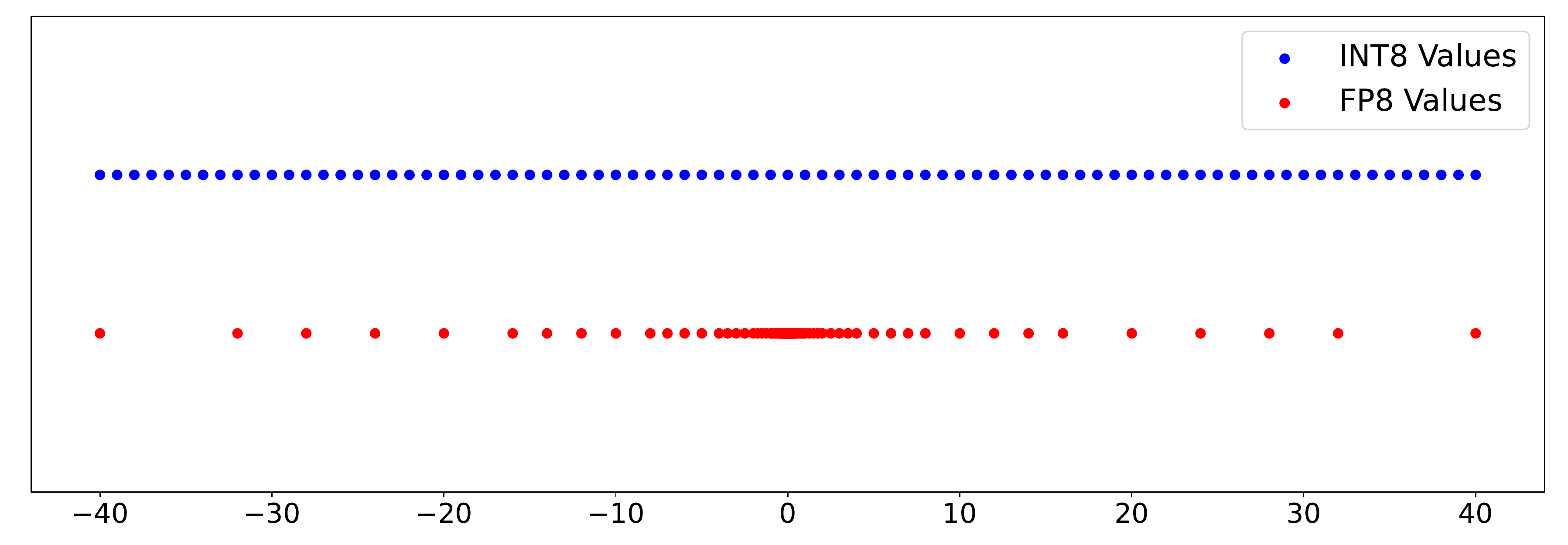}  
    \caption{Value Distribution represented in FP8 and INT8.}  
    \label{fig:distribution}  
  \end{minipage}  
\end{figure}

Mainstream deep learning hardware typically supports high-bit FP and low-bit INT operations, while recently, H100 has introduced support for FP8-E5M2 and FP8-E4M3. 
As for hardware efficiency, FP operations generally incur higher costs and lower performance than INT operations.
However, for low-bit quantization and inference where the bit-width drops to 8/4 bits, their hardware efficiency is not well-established.
Therefore, this work also benchmarks the  hardware cost of INT and FP operations at various bit widths, as will be shown in Section~\ref{sec:ppa}.

\paragraph{Model Quantization}
Model quantization reduces the memory and computational cost of DNNs by using fewer bits to represent tensors.
There are two primary methods of model quantization: weight-only (W-only) quantization and weight-and-activation (WA) quantization.
W-only quantization compresses weight tensors only, still requiring higher bit-width operations during computation. Meanwhile, WA quantization further quantizes activation tensors, enabling lower bit-width computation and improved efficiency. 
Based on whether retraining is required, quantization can be divided into post-training quantization (PTQ) and quantization-aware training (QAT). 
PTQ converts pre-trained models into quantized versions without additional training, making it faster and more cost-effective, while QAT uses a lengthy training process to simulate quantitative effects.
In this work, we focus on applying PTQ to LLMs due to the prohibitive training costs of QAT.

\section{Comparative Analysis of Integer and Floating-Point Formats
}

In this section, we aim to shed light on the differences between INT and FP quantization formats by conducting a comprehensive comparison, specifically focusing on hardware efficiency and quantization error. 
We strive to understand their individual strengths and weaknesses, and how they impact model performance and efficiency. 
To ensure a fair comparison, we maintain the same bit-width for both formats throughout the analysis.

\subsection{Hardware Cost} \label{sec:ppa}

\begin{figure}[htbp]  
    \centering  
    \begin{subfigure}{0.32\textwidth}  
        \includegraphics[width=\textwidth]{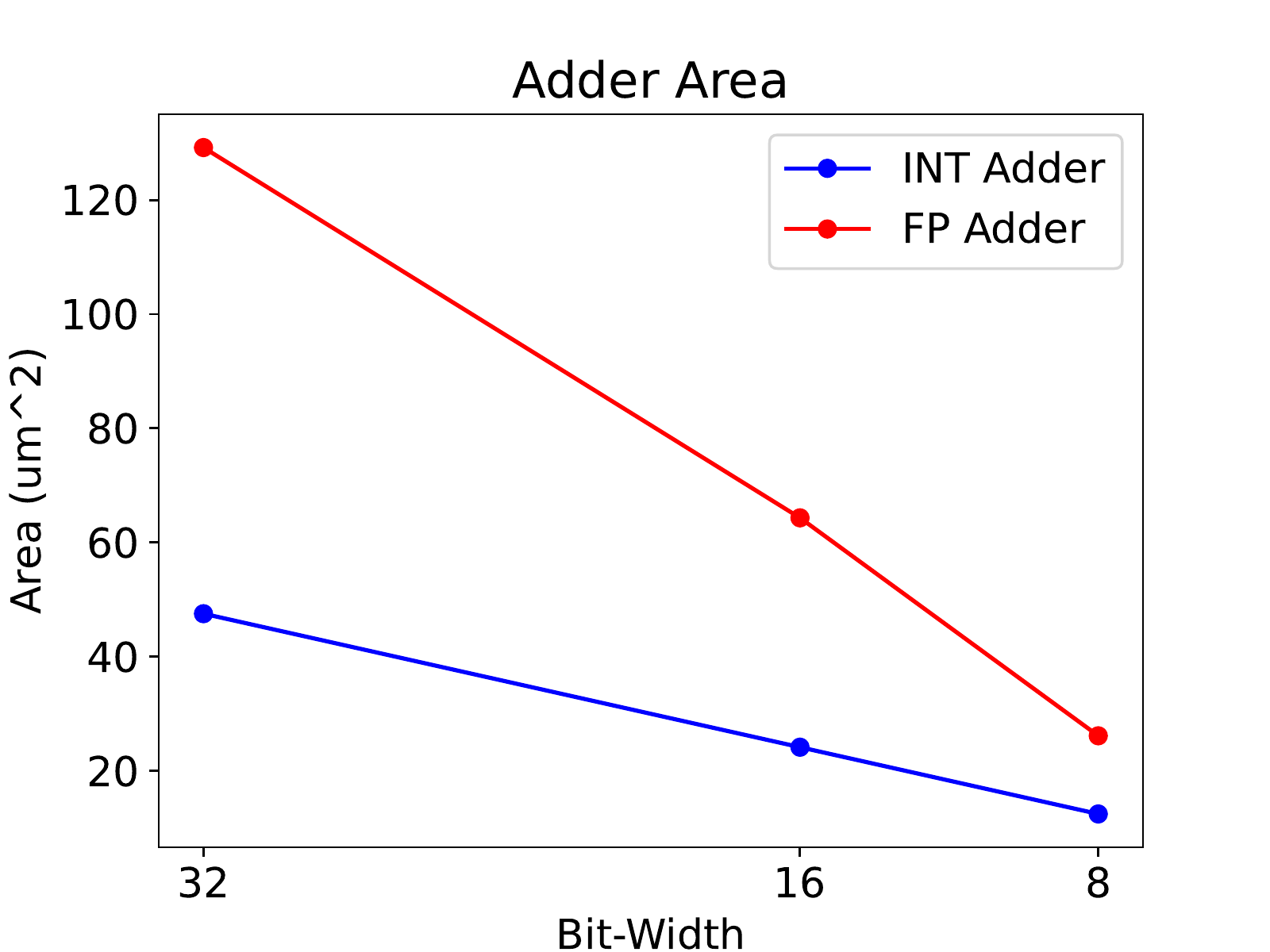}  
    \end{subfigure}  
    \hfill  
    \begin{subfigure}{0.32\textwidth}  
        \includegraphics[width=\textwidth]{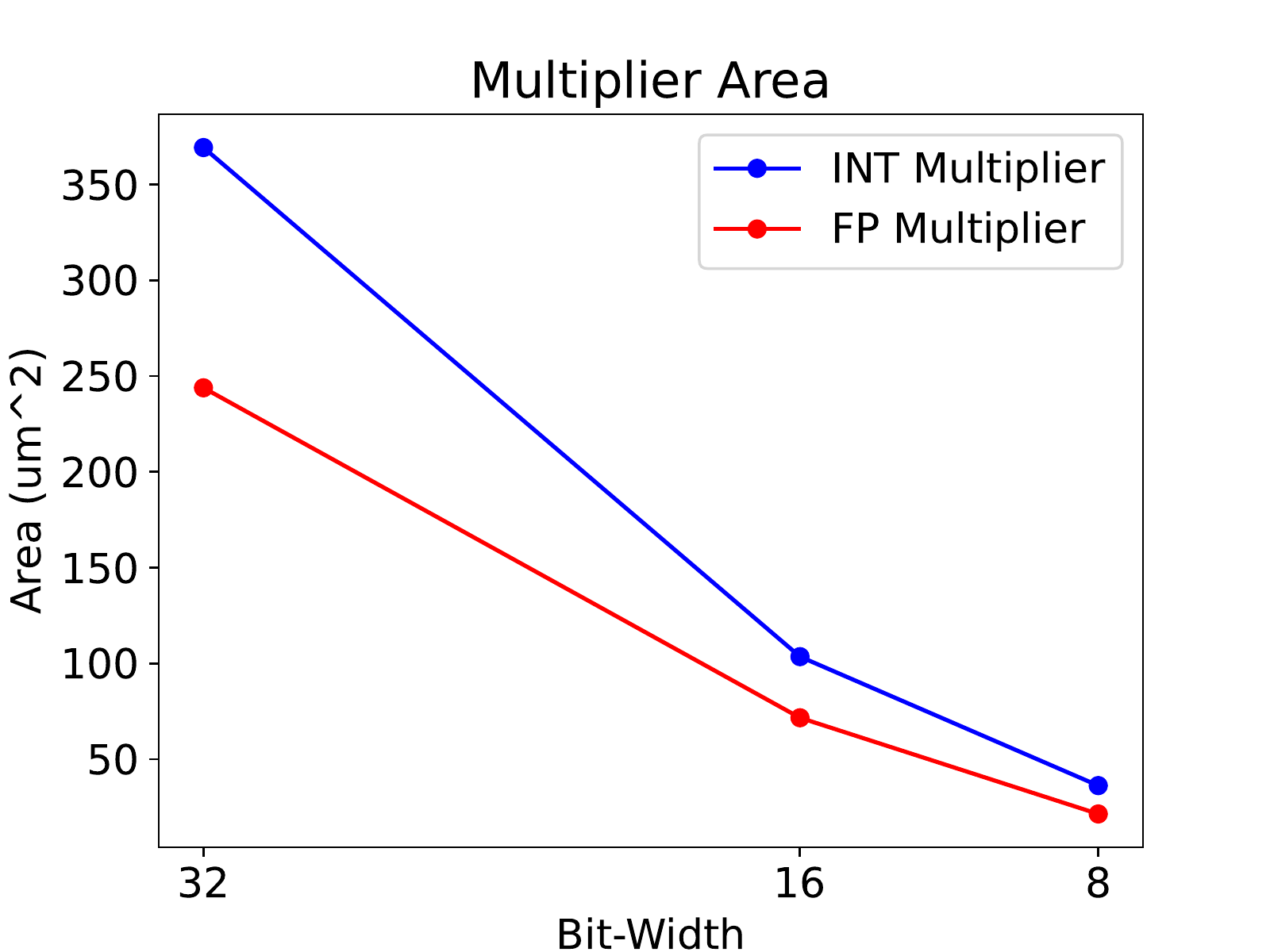}  
    \end{subfigure}  
    \hfill  
    \begin{subfigure}{0.32\textwidth}  
        \includegraphics[width=\textwidth]{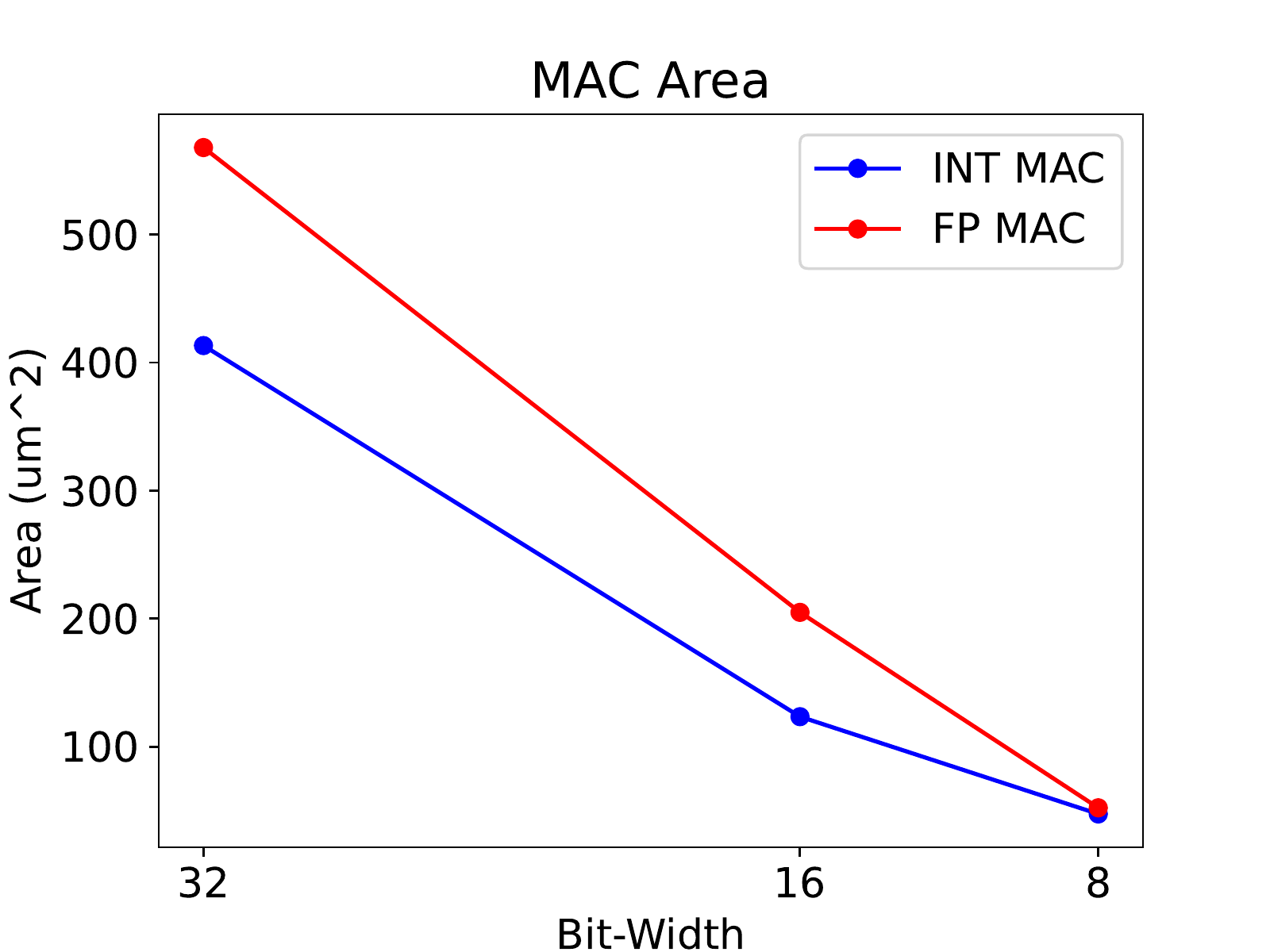}  
    \end{subfigure}  
    \caption{Area differences of INT and FP operators across various bit-widths (32-bit, 16-bit and 8-bit) with TSMC 7nm technology at 0.5GHz. From left to right: Adder, Multiplier, and MAC unit.}  
    \label{fig:area_differences}  
\end{figure}

We first delve into hardware efficiency by comparing the hardware cost of INT and FP operators, including adders, multipliers, and multiply-accumulate (MAC) units, across various bit widths. 
Utilizing Synopsys Design Ware along with TSMC's 7nm technology and the Synopsys Design Compiler, we are able to accurately determine the area requirements for each type of operator. In this experiment, we establish a target clock frequency of 0.5GHz for all operators. 
For 8-bit FP operators, we employ the E5M2 format, while E4M3 yields analogous outcomes. It is worth mentioning that, in 8-bit MAC, the multiplier is 8-bit, and the accumulator is 16-bit to prevent overflow, a standard configuration for low-bit MAC.
As illustrated in Figure~\ref{fig:area_differences}, FP multipliers require less area than their INT counterparts, whereas FP adders necessitate more area than INT adders. Concerning MAC units, which function as a fundamental building block for matrix multiplication in DNNs, FP operations typically demand more area than INT operations.
However, this disparity narrows as the bit-width decreases. Intriguingly, at 8-bit, the area requirements for FP8 and INT8 MAC units are almost identical. This observation indicates that INT8 and FP8 demonstrate similar hardware costs and inference performance, aligning with the specifications of the H100 GPU.

\subsection{Quantization Error}

In this subsection, we compare quantization errors from different formats through statistical analysis of real tensors and layers in LLMs. Quantization primarily targets Linear layers (operators) as they dominate storage and computation. The linear layer is represented as $A_{out} = W * A_{in}$, where $W$ is the weight tensor, $A_{in}$ is the input activation tensor, and $A_{out}$ is the output activation tensor. We analyze quantization errors on all three tensors for a comprehensive understanding. We perform static tensor analysis on $W$, dynamic tensor analysis on $A_{in}$, and layer analysis on $A_{out}$, which is influenced by error accumulation from both inputs.

In our analysis, we use per-channel weight tensor quantization and per-tensor activation quantization, following widely-used approaches in previous quantization research~\cite{gholami2021survey}. The FP4 format adopts an E2M1 configuration, and the FP8 format follows an E4M3 configuration.

\paragraph{Static (Weight) Tensor Analysis}

We quantize weight tensors using both FP and INT formats with 4-bit and 8-bit and calculate the mean squared error (MSE) between quantized and original tensors. 
The weight tensors used in this experiment are all sampled from the LLaMA-65B model~\cite{touvron2023llama}.

\begin{figure}[htbp]  
    \centering  
    \begin{subfigure}{0.45\textwidth}  
        \includegraphics[width=\textwidth]{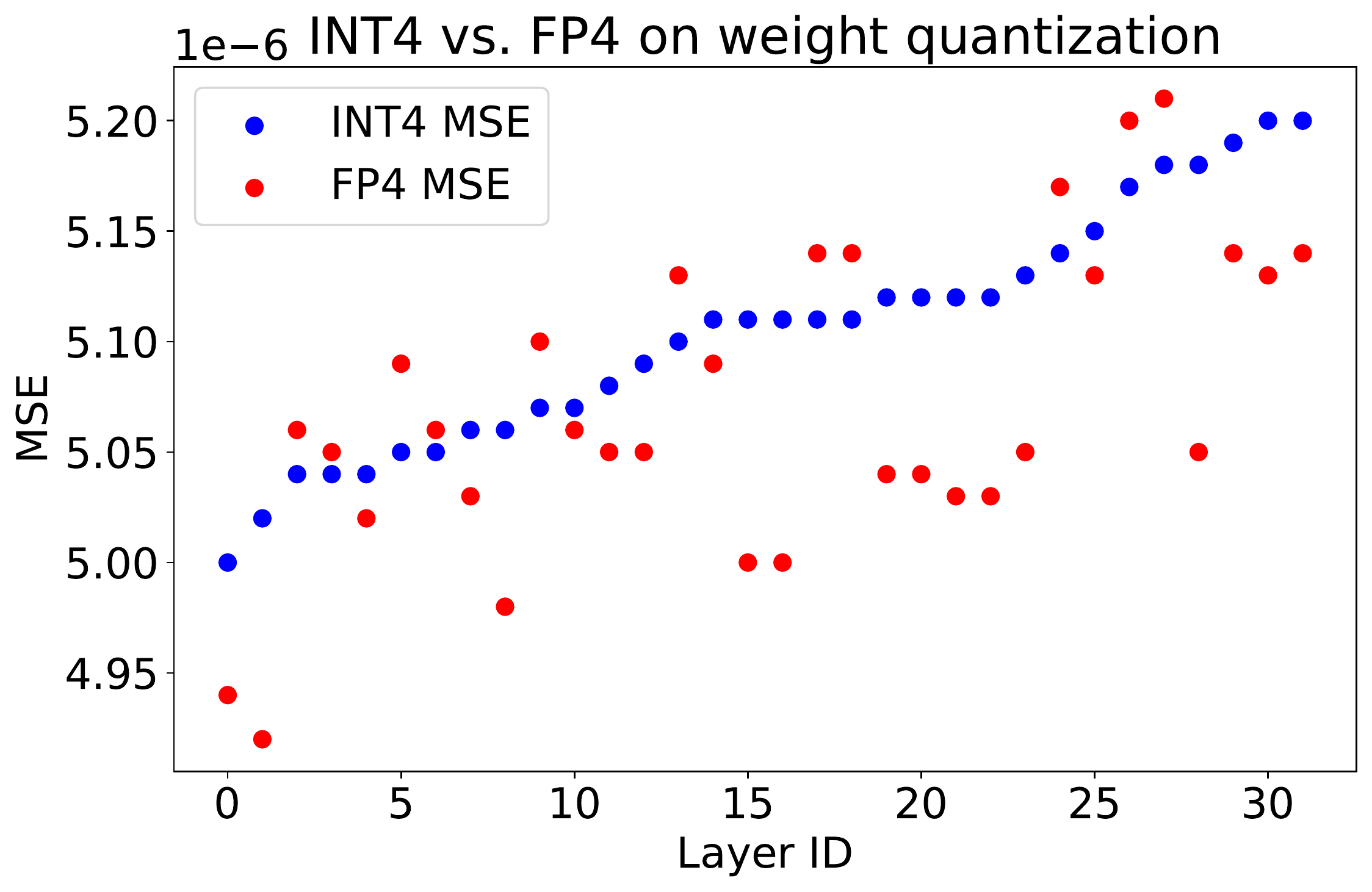}  
    \end{subfigure}  
    \begin{subfigure}{0.45\textwidth}  
        \includegraphics[width=\textwidth]{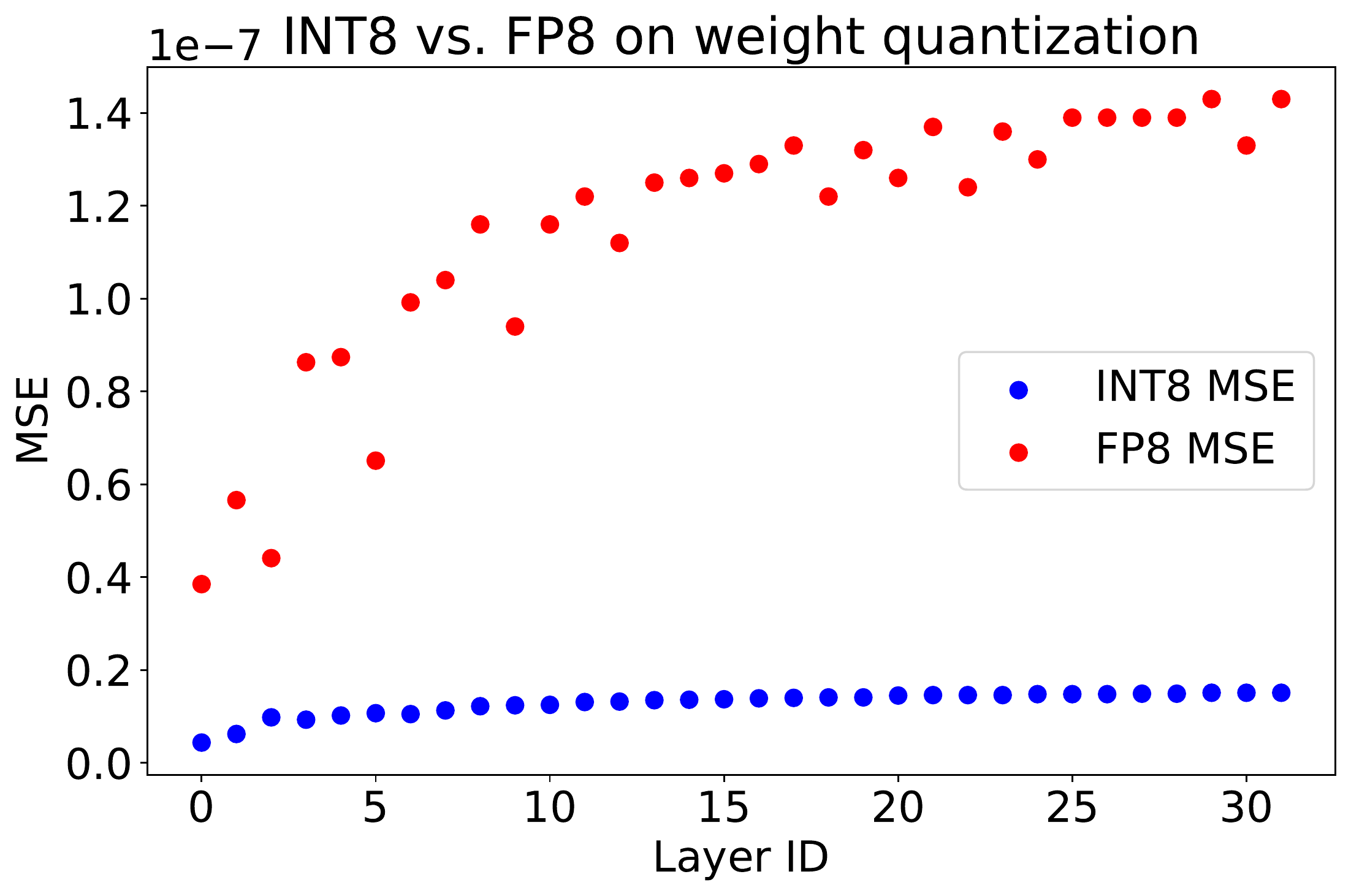}  
    \end{subfigure}  
    \caption{Quantizing weight tensors from various layers of LLaMA-65B with 4-bit (INT4 vs. FP4) and 8-bit (INT8 vs. FP8). No consistent superior format in 4-bit and INT outperforms FP in 8-bit.} 
    \label{fig:staticanalysis} 
\end{figure}

Figure~\ref{fig:staticanalysis} presents the comparison between INT and FP quantization. For clarity, we sort the results by INT MSE. It should be noted that the Layer ID does not correspond directly to the actual number of layers in the model.
When quantizing weight tensors with 8-bit, INT format exhibits a lower error than FP format, as shown in Figure~\ref{fig:staticanalysis}-right. When the quantization bit-width decreases to 4-bit, there is no absolute winner between FP and INT formats as shown in Figure~\ref{fig:staticanalysis}-left. In some layers, FP4 has lower errors, while in others, INT4 has lower errors. 

This analysis result suggests that INT has an advantage in 8-bit weight quantization. However, when the bit-width decreases,  the bits for exponent also decline, making the distribution of INT and FP more similar. The advantage of quantizing static tensors with uniformly distributed INT format fades away in 4-bit quantization, leading to no clear optimal solution for 4-bit quantization of weights.

\begin{figure}[htbp]  
    \centering  
    \begin{subfigure}{0.7\textwidth}  
        \includegraphics[width=\textwidth]{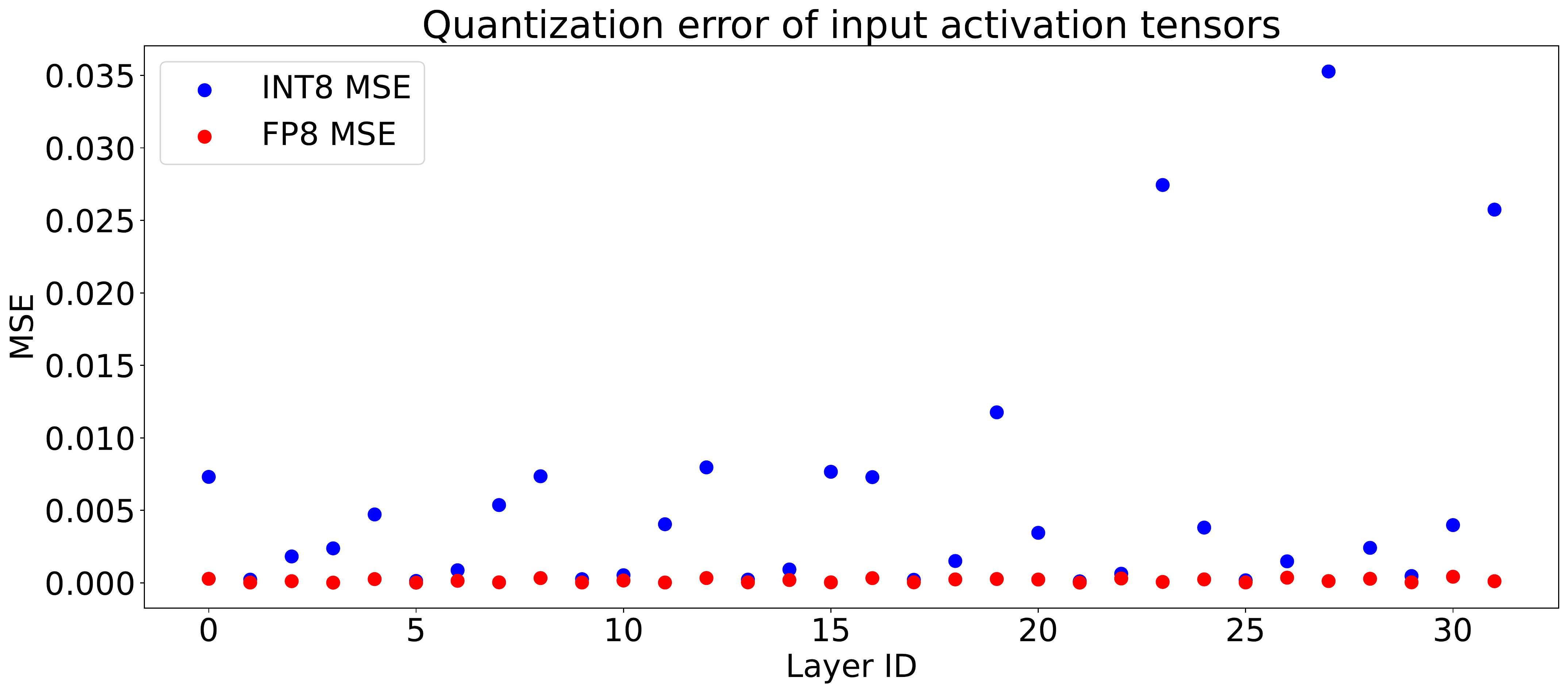}  
    \end{subfigure}  
    \caption{Using scales obtained from different calibration sets to quantize unseen input activation tensors. FP8 exhibits a better adaptability to the scale value than INT8.} 
    \label{fig:dynamicanalysis} 
\end{figure}

\paragraph{Dynamic (Activation) Tensor Analysis}
Unlike static weight tensors during inference, activation tensors are dynamic and change with each input. Therefore, calibration is an essential step in the quantization process for activation tensors to determine the appropriate scaling factors. 

In this analysis, we select 32 layers from the LLaMA-65B model and quantize the input activation tensors of each layer with scales from calibration sets.
Finally, we compare the MSE errors of FP8 and INT8 quantization as shown in Figure\ref{fig:dynamicanalysis}.
Our findings indicate that with scales obtained from calibration sets, the quantization error of FP8 is lower than that of INT8.
This is because the calibration process always selects the largest value among multiple batches to determine the scaling factor in quantization. 
This process tends to derive a larger-than-suitable scaling factor for most of the batches. 
However, the FP format, with lower precision for large values and higher precision for small values, is inherently better suited for quantizing such dynamic tensors.

\paragraph{Layer (Operator) Analysis}

Although quantization error analysis on input tensors can yield some insights, the relationship between the layer output quantization error and the input tensors' quantization errors is not clear.
Therefore, we investigate the quantization errors in the output tensors $A_{out}$ when the input tensors ($W$,$A_{in}$) are quantized using different formats.
$A_{out}$ MSE errors in both W-only (W4A16) and WA (W8A8) quantization scenarios are presented, with $W$ and $A_{in}$ tensors from various layers in the LLaMa-65B model.

\begin{figure}[htbp]  
    \centering  
    \begin{subfigure}{0.32\textwidth}  
        \includegraphics[width=\textwidth]{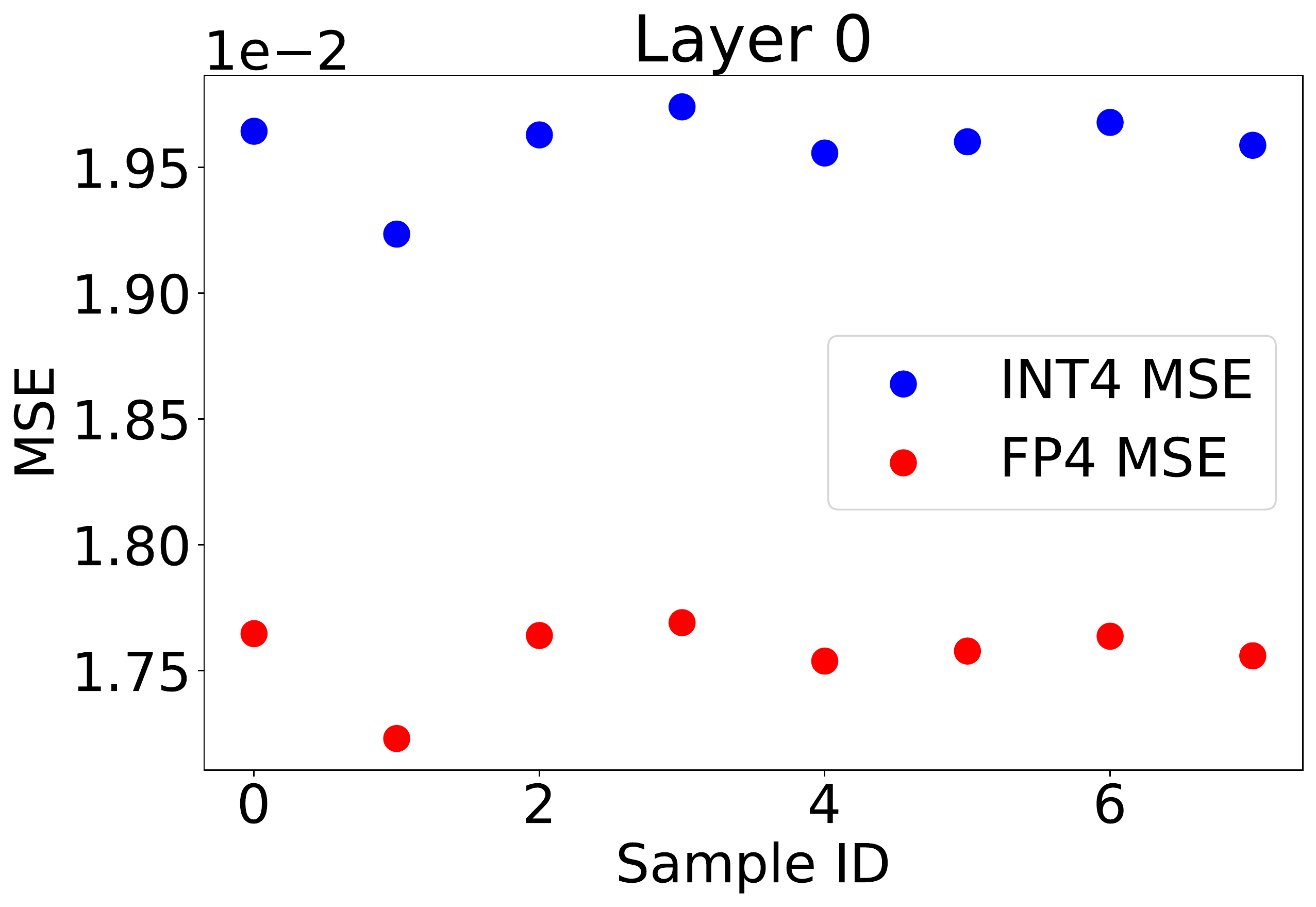}  
    \end{subfigure}  
    \begin{subfigure}{0.31\textwidth}  
        \includegraphics[width=\textwidth]{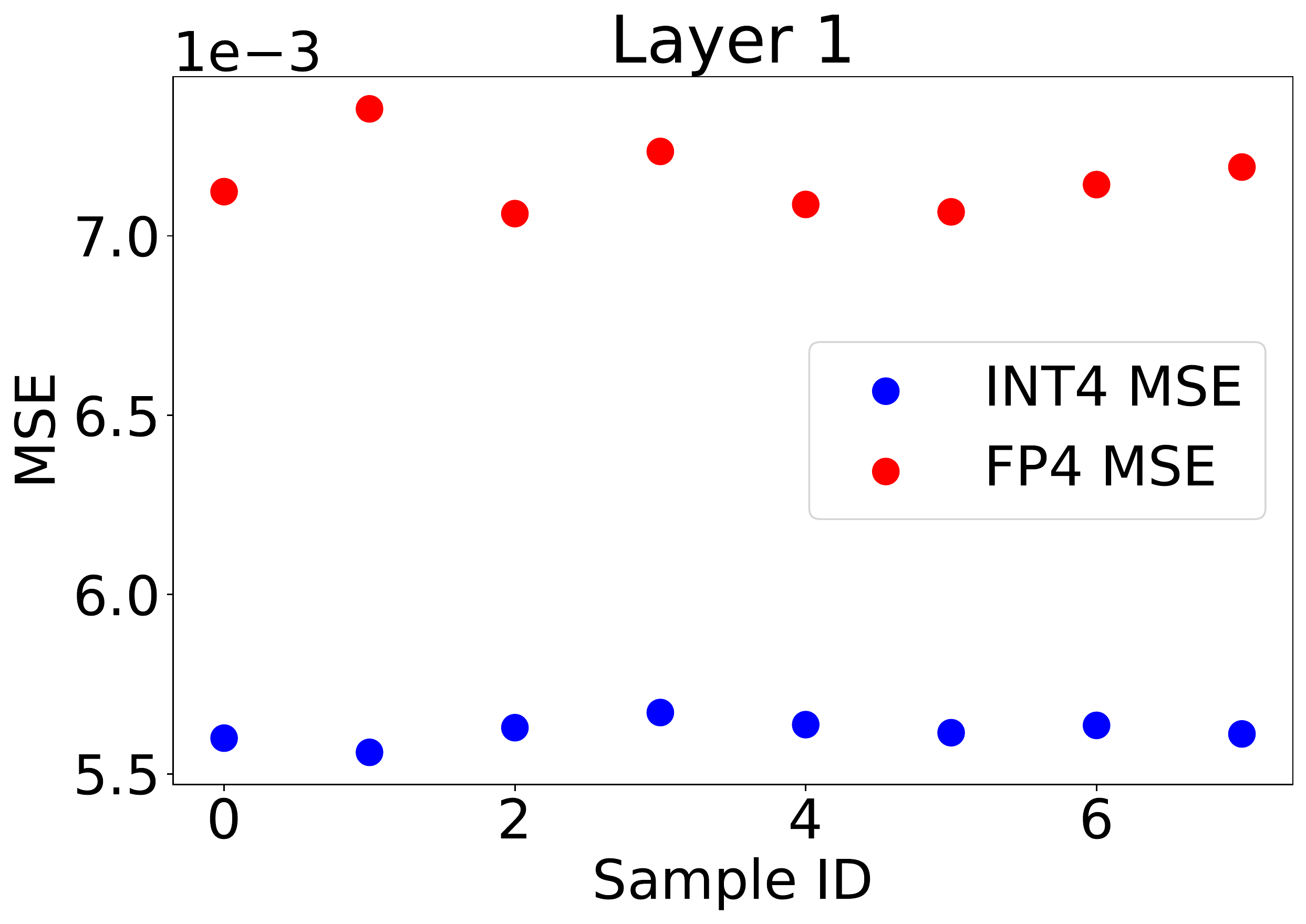}  
    \end{subfigure}  
    \begin{subfigure}{0.32\textwidth}  
        \includegraphics[width=\textwidth]{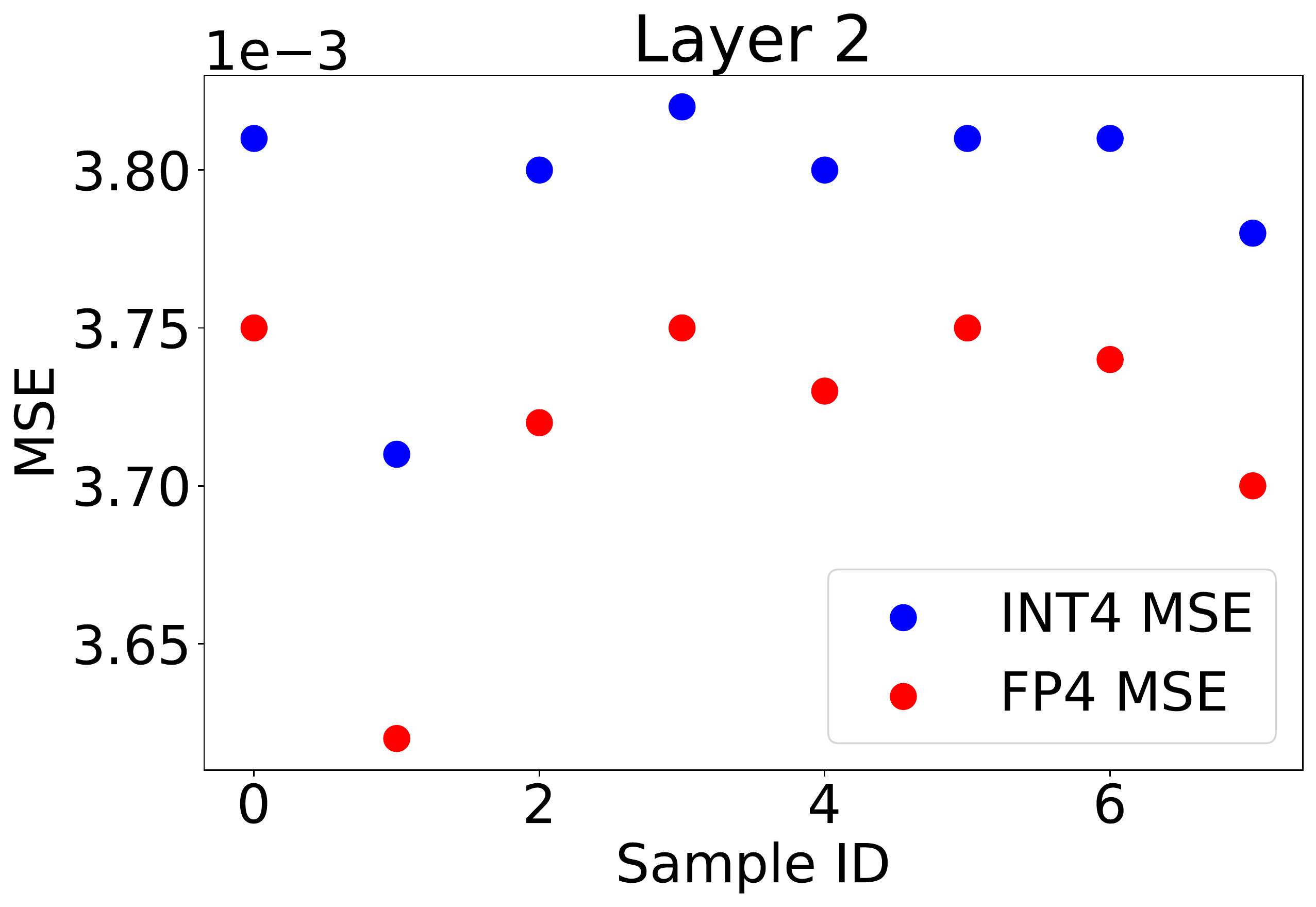}  
    \end{subfigure}
    \caption{Quantization error of output activation tensors across various layers using W4A16 quantization. Different layers exhibit varying preferences for either FP4 or INT4 formats.} 
    \label{fig:W4A16layeranalysis} 
\end{figure}

\begin{figure}[htbp]  
    \centering  
    \begin{subfigure}{0.8\textwidth}  
        \includegraphics[width=\textwidth]{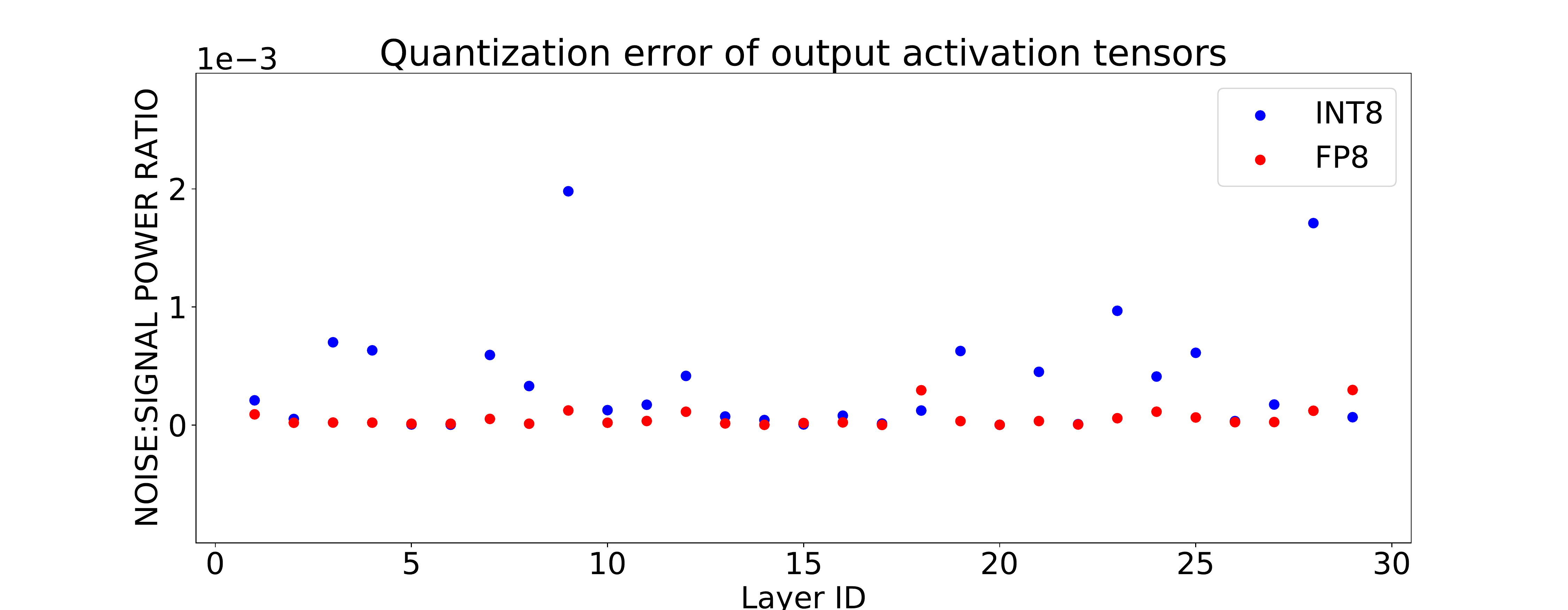}  
    \end{subfigure}  
    \caption{Quantization error of output activation tensors across various layers using W8A8 quantization. Different layers exhibit varying preferences for either FP8 or INT8 formats.} 
    \label{fig:W8A8layeranalysis} 
\end{figure}

Figure~\ref{fig:W4A16layeranalysis} shows the results of W-only quantization with 4-bit. 
As different weight tensors favor different formats between INT4 and FP4, it is natural that different layers exhibit varying preferences between INT4 and FP4. 
Figure \ref{fig:W8A8layeranalysis} shows the results of WA quantization with 8-bit, using the noise-signal power ratio to indicate quantization error (a lower value is preferable). The results suggest that when activation quantization is taken into account, FP8 leads to lower quantization error in most of the layers. However, there are still some cases where INT8 is favored as weight tensors prefer INT8 over FP8.

Based on the above observations, we can conclude that the suitability of FP or INT for LLaMA layers varies on a case-by-case basis when considering layer errors.
Both INT4 and FP4 are useful for weights in different layers, as no optimal 4-bit format exists for static tensor quantization.
Meanwhile, though we have drawn the conclusion that INT8 is better for weights and FP8 is better for activation, the accuracy of $A_{out}$ depends on the impact of $W$ and $A_{in}$ being multiplied, so there is also no consistent superior format for W8A8 quantization.
In Section~\ref{sec:exp}, we will demonstrate that an appropriate format choice for each layer can result in better model accuracy compared to using the same format for all layers.

\section{Mixture of Formats Quantization}
\subsection{Exploiting the Complementary Advantages of Integer and Floating Point Formats}

Given the analysis findings that no single quantization format consistently outperforms the others in all scenarios, we propose the \textbf{Mixture-of-Formats Quantization (MoFQ)} method. The key idea behind MoFQ is to leverage the complementary advantages of integer (INT) and floating-point (FP) formats in a unified framework, thereby maximizing the potential benefits of both formats.

\begin{algorithm} [h] 
\caption{Mixture-of-Formats Quantization (MoFQ)}
\label{alg:mofq}
\begin{algorithmic}[1]  
\Function{MoFQ}{$model, is\_w\_only, format\_candidates, bit\_width, error\_metric$}  
    \State $quantized\_model \gets \Call{initialize\_empty\_model}{}$  
    \For{$layer$ in $model.layers\_to\_quant$}  
        \State $min\_error \gets \infty, best\_format \gets None$  
        \For{$format$ in $format\_candidates$}  
            \State $quantized\_layer \gets \Call{quantize\_layer}{layer, format, bit\_width, is\_w\_only}$  
            \State $error \gets \Call{compute\_error}{layer, quantized\_layer, error\_metric}$  
            \If{$error < min\_error$}  
                \State $min\_error \gets error, \ best\_format \gets format$  
            \EndIf  
        \EndFor  
        \State $quantized\_layer \gets \Call{quantize\_layer}{layer, best\_format, bit\_width, is\_w\_only}$  
        \State $\Call{add\_layer\_to\_model}{quantized\_model, quantized\_layer}$  
    \EndFor  
    \State \Return $quantized\_model$  
\EndFunction  
  
\Function{quantize\_layer}{$layer, format, bit\_width, is\_w\_only$}  
    \State \Comment{Quantize the given layer with the specified format, bit\_width, and is\_w\_only flag}  
\EndFunction

\Function{compute\_error}{$original\_layer, quantized\_layer, error\_metric$}  
    \State \Comment{Calculate the error between original and quantized layers using the error\_metric}  
\EndFunction  
\end{algorithmic}  
\end{algorithm}

Specifically, MoFQ allows for the selection of the most suitable format on a layer-by-layer basis. As illustrated  in Algorithm~\ref{alg:mofq}, the algorithm considers the model to be quantized, a flag \texttt{is\_w\_only}, format candidates (e.g., INT and FP), bit width (e.g., 8 or 4), and error metric as inputs. 
If \texttt{is\_w\_only} is set to true, only the weight tensor in the layer will be quantized (i.e., W-only quantization); otherwise, both weights and activations will be quantized (i.e., WA quantization), resulting in a quantized operator that can leverage low-bit hardware matrix multiplication units.
We concentrate on layer-wise format selection with the same format for each tensor/layer and the same bit-width across all quantized tensors/layers, as this approach adheres to the simplicity principle, offering a straightforward and efficient implementation with minimal adaptation to the existing system or hardware. 
Such a straightforward method can achieve satisfactory results, as will be demonstrated in Section~\ref{sec:exp}.
Utilizing a finer granularity than the layer level or increasing the bit-width can indeed improve the quantization effect, but it also comes at the cost of increased memory consumption and higher system or hardware complexity. 
We leave the exploration of this trade-off as future work, as it requires further investigation and analysis.

The selection of the error metric plays a crucial role in MoFQ, as it influences the balance between quantization accuracy and speed. Several metrics are available, including tensor error, layer output error, and model output error, from less precise to more precise. A more precise metric may lead to better quantization results but at the expense of increased computational time. Thus, choosing an appropriate error metric is essential for achieving the desired balance between accuracy and efficiency in the quantization process. 
It's worth mentioning that the evaluation metric for determining the superiority of a format is empirical and doesn't have an absolute standard. For example, a lower tensor error may indicate a higher likelihood of achieving better model accuracy, but it isn't guaranteed. In MoFQ, we use various selection metrics to find the right balance between quantization accuracy and speed. Our empirical observations suggest that using tensor error or layer output error suffices to guide the format selection for W-only quantization. Meanwhile, for WA quantization, the model output error offers the best format selection choices.

\subsection{Reallocating NaN and Inf for Enhanced Low-Bit Floating Point Quantization}

In low-bit quantization, maximizing number representation efficiency is essential for achieving the best possible precision and expressiveness. One optimization opportunity lies in reallocating the special NaN (Not a Number) and Inf (Infinity) values in standard floating-point formats.
In IEEE floating-point formats, an exponent field with all bits set to 1 denotes NaN and Inf values. However, during model quantization, these special values serve no practical purpose, resulting in wasted bit combinations that could otherwise be used to represent a broader range of numbers. By reallocating NaN and Inf representations to normalized numbers, we can enhance the precision and expressiveness of the low-bit FP format for improved model quantization.

In practical quantization scenarios, the impact of NaN and Inf redundancy varies depending on the number of bits used for representation. For instance, in an 8-bit floating-point format (FP8), the impact is relatively minor as it can represent 256 numbers, with NaN and Inf occupying only a small portion. 
On the other hand, FP8 is primarily used for WA quantization, which requires hardware support for matrix multiplication. Therefore, it may not be suitable to modify FP8 due to potential hardware compatibility issues.
However, in a 4-bit format (FP4), the impact becomes more significant  as it can only represent 16 numbers in total. As FP4 is primarily used for W-only quantization, format modification can be addressed at the software level.
As illustrated in Table~\ref{tab:fp4}, 4 of these numbers are used to represent NaN and Inf when adhering to the IEEE 754 standard. By reallocating them, we can obtain additional represented numbers, specifically \(\pm4\) and \(\pm6\). Tensor-wise analysis shows that our redesigned FP4 format can lead to about 35\% lower quantization errors compared to the IEEE-aligned FP4 format. This improvement makes more layers prefer our re-designed FP4 than INT4 in W-only quantization.

\begin{table}[ht]  
\caption{Numbers represented in FP4-E2M1 with NaN and Inf (IEEE 754 standard) and Numbers represented in FP4-E2M1 without NaN and Inf (Our design).} 
\centering  
\resizebox{\columnwidth}{!}{%
\footnotesize  
\begin{tabular}{@{}c|*{16}{c}@{}}  
\toprule  
UINT4 & 0 & 1 & 2 & 3 & 4 & 5 & 6 & 7 & 8 & 9 & 10 & 11 & 12 & 13 & 14 & 15 \\  
\midrule  
FP4 w/ NaN\&Inf & 0 & 0.5 & 1 & 1.5 & 2 & 3 & Inf & NaN & -0 & -0.5 & -1 & -1.5 & -2 & -3 & Inf & NaN \\  
FP4 w/o NaN\&Inf & 0 & 0.5 & 1 & 1.5 & 2 & 3 & 4 & 6 & -0 & -0.5 & -1 & -1.5 & -2 & -3 & -4 & -6 \\  
\bottomrule  
\end{tabular}%
}    
\label{tab:fp4}  
\end{table}

\section{Experiments} \label{sec:exp}

In the experiments, we present the validation results of our MoFQ approach for both W-only (W4A16) and WA (W8A8) quantization scenarios separately. We apply per-channel quantizations to weight tensors and per-tensor quantizations to activation tensors. The FP4 construction excludes NaN and Inf, using E2M1, while FP8 construction employs E4M3.
For validation, we utilize LLaMA\cite{touvron2023llama} and OPT\cite{zhang2022opt} models. Validation datasets include: 1) WikiText-2\cite{merity2016wikitext}; 2) LAMBADA\cite{paperno2016lambada}; 3) PIQA\cite{bisk2020piqa}; 4) HellaSwag\cite{zellers2019hellaswag}. Our MoFQ implementation is based on the PPQ library\cite{ppq} and GPTQ\cite{gptq}.

\subsection{W-only 4bit quantization with SOTA quantization errors}

Table~\ref{tab:W4Wiki} compares various quantization methods on LLaMA models, including INT4(GPTQ), FP4(ours), and MoFQ4. INT4(GPTQ) is the SOTA method from the GPTQ paper\cite{frantar2022gptq}, FP4(ours) is our FP4 format, and MoFQ4 is mixture of FP4(ours) and INT4 formats. Evaluation results show that FP4(ours) and MoFQ4 generally outperform INT4(GPTQ), with MoFQ4 often yielding better results than FP4(ours). However, MoFQ doesn't consistently surpass FP4, indicating room for further improvement in our MoFQ approach.

\begin{table}[H]
\caption{Weight-only quantization results On WikiText-2, LAMBADA, PIQA and HellaSwag datasets. For WikiText-2 dataset, we show perplexity metric. For the other three, we show average accuracy.}
\resizebox{\textwidth}{!}{%
\begin{tabular}{c|cccc|cccc}
\hline
 & \multicolumn{4}{c|}{WikiText-2 $\downarrow$} & \multicolumn{4}{c}{LAMBADA $\uparrow$} \\ \hline
 & FP16 &  \begin{tabular}[c]{@{}c@{}}INT4\\ (GPTQ)\end{tabular} & \begin{tabular}[c]{@{}c@{}}\textbf{FP4}\\ \textbf{(ours)}\end{tabular} & \begin{tabular}[c]{@{}c@{}}\textbf{MoFQ4}\\ \textbf{(FP\%)}\end{tabular} & FP16  & \begin{tabular}[c]{@{}c@{}}INT4\\ (GPTQ)\end{tabular} & \begin{tabular}[c]{@{}c@{}}\textbf{FP4}\\ \textbf{(ours)}\end{tabular} & \begin{tabular}[c]{@{}c@{}}\textbf{MoFQ4}\\ \textbf{(FP\%)}\end{tabular} \\\hline
LLaMA-7B & 5.68 &  6.38 & 6.04 & \textbf{6.03(88.9\%)} & 0.884 &  0.862 & 0.878 & \textbf{0.878(88.9\%)} \\
LLaMA-13B &  5.09 & 5.40 & 5.35 & \textbf{5.33(97.1\%)} & 0.883 &  0.877 & \textbf{0.879} & 0.874(97.1\%) \\
LLaMA-33B &  4.10 & 4.36 & 4.33 & \textbf{4.30(97.9\%)} & 0.862 &  0.853 & \textbf{0.858} & 0.845(97.9\%) \\
LLaMA-65B &  3.53 & 3.85 & 3.85 & \textbf{3.78(96.8\%)} & 0.909 &  0.907 & 0.911 & \textbf{0.916(96.8\%)} \\ \hline
 & \multicolumn{4}{c|}{PIQA $\uparrow$} & \multicolumn{4}{c}{HellaSwag $\uparrow$} \\ \hline
 & FP16 & \begin{tabular}[c]{@{}c@{}}INT4\\ (GPTQ)\end{tabular} & \begin{tabular}[c]{@{}c@{}}\textbf{FP4}\\ \textbf{(ours)}\end{tabular} & \begin{tabular}[c]{@{}c@{}}\textbf{MoFQ4}\\ \textbf{(FP\%)}\end{tabular} & FP16  & \begin{tabular}[c]{@{}c@{}}INT4\\ (GPTQ)\end{tabular} & \begin{tabular}[c]{@{}c@{}}\textbf{FP4}\\ \textbf{(ours)}\end{tabular} & \begin{tabular}[c]{@{}c@{}}\textbf{MoFQ4}\\ \textbf{(FP\%)}\end{tabular} \\\hline
LLaMA-7B & 0.780 &  0.764 & \textbf{0.781} & 0.776(88.9\%) & 0.558  & 0.476 & \textbf{0.519} & 0.517(88.9\%) \\
LLaMA-13B & 0.783  & 0.792 & 0.806 & \textbf{0.808(97.1\%)} & 0.587  & \textbf{0.564} & 0.560 & 0.562(97.1\%) \\
LLaMA-33B & 0.787  & \textbf{0.788} & 0.781 & 0.784(97.9\%) & 0.605  & 0.582 & \textbf{0.585} & 0.582(97.9\%) \\
LLaMA-65B & 0.769  & 0.761 & 0.761 & \textbf{0.764(96.8\%)} & 0.551  & 0.539 & 0.543 & \textbf{0.544(96.8\%)} \\ \hline

\end{tabular}%
}
\label{tab:W4Wiki}
\end{table}

\begin{table}[H]
\centering
\caption{The  quantization time(s) of INT4(GPTQ), FP4(ours) and MoFQ4 on LLaMA models. The speedup over INT4(GPTQ) is shown in brackets.}
\begin{tabular}{|c|c|c|c|c|}
\hline
 & LLaMA-7B & LLaMA-13B & LLaMA-33B & LLaMA-65B \\ \hline
INT4(GPTQ) & 389 & 1088 & 2535 & 4684 \\ \hline
\textbf{FP4(ours)} & 5(77.8x) & 9(120.8x) & 19(133.4x) & 36(130.1x) \\ \hline
\textbf{MoFQ4} & 42(9.3x) & 69(15.8x) & 162(15.6x) & 319(14.7x) \\ \hline
\end{tabular}
\label{tab:speedup}  
\end{table}

In addition to comparing quantization accuracy, we also examine the runtime required for quantizing the models. 
As shown in Table~\ref{tab:speedup}, we observe that the quantization speed of FP4 is more than 77.8 times faster than INT4(GPTQ) (with calibration nsample = 128), while MoFQ4 is over 9.3 times faster compared to INT4(GPTQ). 
This can be attributed to the fact that GPTQ necessitates real-time updates of the original weight tensors during the quantization process, resulting in considerable computational overhead. 
In contrast, FP4(ours) and MoFQ4 adopt the naive linear quantization with round-to-nearest (RTN), leading to significantly faster speeds.

\subsection{W8A8 quantization with accuracy close to full-precision models}

In this subsection, we compare the performance of various quantization methods on LLaMA and OPT models. The methods evaluated include INT8, FP8, and MoFQ8, where MoFQ8 represents the mixture of FP8 and INT8 formats in the quantization method.

Table~\ref{tab:W8A8piqa} compares various quantization methods on four datasets: WikiText-2, LAMBADA, PIQA, and HellaSwag. FP8 consistently outperforms INT8 across all cases. This can be attributed to the increased challenge of quantizing dynamic activation tensors compared to static weight tensors, as dynamic tensors exhibit more varied distributions and larger outlier values\cite{xiao2022smoothquant}. In Section 3, we determine that FP8 is better suited for quantizing dynamic tensors than INT8. Consequently, despite FP8's potentially weaker performance on weight tensors compared to INT8, it still manages to achieve lower overall quantization errors for the models.

\begin{table}[H]
\caption{WA-Quantization results On WikiText-2, LAMBADA, PIQA and HellaSwag datasets. For WikiText-2 dataset, we show perplexity metric. For the other three, we show average accuracy.}
\resizebox{\textwidth}{!}{%
\begin{tabular}{c|cccc|cccc}
\hline
    & \multicolumn{4}{c|}{WikiText-2 $\downarrow$} & \multicolumn{4}{c}{LAMBADA $\uparrow$} \\ \hline
    & FP16 &  \begin{tabular}[c]{@{}c@{}} INT8\end{tabular} & \begin{tabular}[c]{@{}c@{}}FP8\end{tabular} & \begin{tabular}[c]{@{}c@{}}MoFQ8\\ (FP\%)\end{tabular} 
    & FP16 &  \begin{tabular}[c]{@{}c@{}} INT8\end{tabular} & \begin{tabular}[c]{@{}c@{}}FP8\end{tabular} & \begin{tabular}[c]{@{}c@{}}MoFQ8\\ (FP\%)\end{tabular}  \\
    \hline
LLaMA-7B & 5.68 &  368.21 & 6.59 & \textbf{6.49(87.2\%)} & 0.884 &  0.010 & 0.851 & \textbf{0.887(82.0\%)} \\
LLaMA-13B &  5.09 & 637.95 & 5.64 & \textbf{5.41(86.1\%)} & 0.883 &  0.230 & 0.854 & \textbf{0.881(83.0\%)} \\
LLaMA-33B &  4.10 & 10069.14 & 5.38 & \textbf{5.31(92.7\%)} & 0.862 &  0.000 & 0.822 & \textbf{0.859(90.0\%)} \\ \hline

OPT-350M & 23.27 &  432.86 & 24.46 & \textbf{23.64(71.8\%)} & 0.674 &  0.290 & 0.658 & \textbf{0.669(69.2\%)} \\
OPT-1.3B &  15.44 & 37.72 & 16.78 & \textbf{16.07(78.8\%)} & 0.758 &  0.716 & 0.735 & \textbf{0.746(80.0\%)} \\
OPT-2.7B &  13.08 & 27.56 & 14.24 & \textbf{13.25(83.3\%)} & 0.778 &  0.693 & 0.764 & \textbf{0.777(80.0\%)} \\
OPT-6.7B &  11.43 & 964.58 & 12.41 & \textbf{11.68(89.1\%)} & 0.806 &  0.164 & 0.762 & \textbf{0.800(89.4\%)} \\ 
OPT-13B &  10.68 & 11858.78 & 12.52 & \textbf{10.79(87.2\%)} & 0.802 &  0.001 & 0.724 & \textbf{0.801(84.1\%)} \\
OPT-30B &  10.09 & 13195.34 & 10.95 & \textbf{10.17(89.4\%)} & 0.813 &  0.007 & 0.744 & \textbf{0.812(86.0\%)} \\

\hline
    & \multicolumn{4}{c|}{PIQA $\uparrow$} & \multicolumn{4}{c}{HellaSwag $\uparrow$} \\ \hline
    & FP16 &  \begin{tabular}[c]{@{}c@{}} INT8\end{tabular} & \begin{tabular}[c]{@{}c@{}}FP8\end{tabular} & \begin{tabular}[c]{@{}c@{}}MoFQ8\\ (FP\%)\end{tabular} 
    & FP16 &  \begin{tabular}[c]{@{}c@{}} INT8\end{tabular} & \begin{tabular}[c]{@{}c@{}}FP8\end{tabular} & \begin{tabular}[c]{@{}c@{}}MoFQ8\\ (FP\%)\end{tabular}  \\
    \hline
LLaMA-7B & 0.780 &  0.539 & 0.706 & \textbf{0.779(85.8\%)} & 0.558  & 0.258 & 0.524 & \textbf{0.560(80.6\%)} \\
LLaMA-13B & 0.783  & 0.532 & 0.768 & \textbf{0.788(82.3\%)} & 0.587  & 0.264 & 0.576 & \textbf{0.585(81.7\%)} \\
LLaMA-33B & 0.787  & 0.530 & 0.767 & \textbf{0.789(81.3\%)} &  0.605 & 0.260 & 0.599 & \textbf{0.604(89.1\%)}\\ \hline

OPT-350M & 0.619 &  0.554 & 0.615 & \textbf{0.621(71.9\%)} & 0.292 &  0.265 & \textbf{0.295} & 0.293(73.4\%) \\
OPT-1.3B &  0.693 & 0.667 & 0.690 & \textbf{0.691(75.2\%)} & 0.351 &  0.351 & 0.351 & \textbf{0.351(80.8\%)} \\
OPT-2.7B &  0.708 & 0.687 & 0.712 & \textbf{0.714(75.1\%)} & 0.379 &  0.383 & 0.393 & \textbf{0.396(80.3\%)} \\ 
OPT-6.7B &  0.721 & 0.609 & 0.718 & \textbf{0.720(87.0\%)} & 0.409 &  0.279 & 0.407 & \textbf{0.411(84.4\%)} \\ 
OPT-13B &  0.716 & 0.516 & 0.688 & \textbf{0.715(82.9\%)} & 0.421 &  0.263 & 0.417 & \textbf{0.426(85.0\%)} \\
OPT-30B &  0.725 & 0.524 & 0.713 & \textbf{0.727(80.5\%)} & 0.442 &  0.260 & 0.433 & \textbf{0.442(85.5\%)} \\
\hline
\end{tabular}%
}
\label{tab:W8A8piqa}  
\end{table}

Crucially, we find that under our MoFQ8 quantization method, the accuracy of the quantized model remains remarkably close to that of the FP16 model, irrespective of the model size or dataset. 
This suggests that MoFQ8 effectively chooses the most appropriate format (INT8 or FP8) for each layer's distribution, ultimately achieving W8A8 quantization results with minimal accuracy loss.

\section{Related Work}
Large Language Models (LLMs) have significantly transformed the field of natural language processing, introducing new challenges and opportunities in model quantization.
ZeroQuant~\cite{yao2022zeroquant} and nuQmm~\cite{park2022nuqmm} employ finer (group-wise) granularities to quantize tensors, which requires customized CUDA kernels. 
LLM.int8()~\cite{dettmers2022llm} uses mixed precision (INT8+FP16) to quantize individual tensors or layers in LLMs. However, this approach results in substantial latency overhead. While MoFQ maintains the same data type and bit-width for each tensor/layer. SmoothQuant~\cite{xiao2022smoothquant} improves quantization accuracy on LLMs by offline migrating the quantization difficulty from activations to weights. While previous research has primarily focused on low-bit integer quantization, MoFQ incorporates low-bit floating points (FP8 and FP4) for LLM quantization.

\section{Conclusion}
In this paper, we extensively investigate and compare low-bit integer (INT) and floating-point (FP) formats for quantizing LLMs.
Our findings reveal that due to the complexity and diversity of tensor distribution, the optimal quantization format varies across different layers.
Therefore, we propose the Mixture of Formats Quantization (MoFQ) approach, which selectively determines the optimal format from INT and FP with the same bit-width on a layer-wise basis.
MoFQ is simple, effective, and efficient in format selection and model performance, achieving state-of-the-art results in both W-only and WA quantization.

Despite the promising results demonstrated by MoFQ, there are certain limitations and opportunities for future work: 
1) The analysis presented in this paper is primarily empirical. Although this provides valuable insights into the practical performance of the proposed MoFQ, additional theoretical investigation is necessary to gain a deeper understanding of the principles and mechanisms involved in different quantization formats. 
2) MoFQ can be extended to finer granularities, such as channel-wise or block-wise selection of optimal formats, to further enhance model accuracy.

\bibliographystyle{plain}
\bibliography{reference}

\end{document}